\documentclass[12pt]{article}

\usepackage{amssymb}
\usepackage{amsfonts}
\usepackage{graphicx}
\usepackage{amsmath}
\usepackage{subfigure}

\begin{document}

\title{A multilateral filtering method applied to airport runway image}

\author{ZHANG Yu \thanks{zhagyu@163.com}, SHI Zhong-ke, WANG Run-quan\\
Air Traffic Management System Institute,\\
Northwestern Polytechnical University, Xi'an, 710072, P. R. China\\}

\date{}
\maketitle

\begin{abstract}

By considering the features of the airport runway image filtering,
an improved bilateral filtering method was proposed which can remove
noise with edge preserving. Firstly the steerable filtering
decomposition is used to calculate the sub-band parameters of 4
orients, and the texture feature matrix is then obtained from the
sub-band local median energy. The texture similar, the spatial
closer and the color similar functions are used to filter the
image.The effect of the weighting function parameters is
qualitatively analyzed also. In contrast with the standard bilateral
filter and the simulation results for the real airport runway image
show that the multilateral filtering is more effective than the
standard bilateral filtering.


\end{abstract}

\section{Introduction}

\noindent

The vision based UAV autonomous landing problem receive considerable
amount of attention in recent years \cite{SMS1, SMS2, Hintze},
especially because of the advantage of low load and risk, shorten
the time of re-takeoff, that fixed wing UAV autonomous landing on
airport runway becomes one of the key subjects. The key of this
problem is accurately recognition of runway in order to obtain
information needed by landing. But because there are a lot of
inevitable noises in adopted image sequences, as well as some
inherent disturb such as the mark on runway and plane shadow will
effect accuracy of runway recognition, even cause misleading or
runway recognition failing, so the filtering is necessary.

The Denoise filter will cause an edge blurring problem inevitably
and effect later image processing. In order to segment and recognize
the airport runway from the background and provide useful
information for the UAV landing, the edge preserving in filtered
image is very necessary. The Bilateral filtering (BF)  \cite{TM,
Barash}has predominance of smoothing image with edges preserved. BF
rely on spatial and range differences. When both of them are very
small, BF deduces to the average filtering and the image edge will
be blurred. For instance, the grasslands around the airport become
yellow in autumn and winter or some highlight disturbs exist, the
color of the runway is similar with the background. In this
situation, BF is not suitable for solve this problem. While now
there still isn't any correlative research in airport runway
filtering problems. To solve this problem, with the consideration of
the realtime requirement of the system, it is necessary to introduce
more features to improve the robustness and effects of the filtering
method. Focus on the reality that the colors of the runway and the
background are simillar, in this paper we used an improved bilateral
filter in the airport image filtering and compared the simulation
results, researched the real runway image experimently.

This paper is organized as follows. In next section we first give a
simple introduction of the standard bilateral filtering theory. Then
in Section III the improved filtering and texture extraction method
were developed. The simulation results and application in airport
image processing is provided in section IV. Some conclusions
suggestions are given in the final section.

\section{Bilateral filter}

\noindent

Given an input image $I \left ( {\bf x} \right )$, the bilateral
filtering output image $I' \left ( {\bf x} \right )$ can be received
as follow:
\begin{equation}
I' \left ( {\bf x} \right ) = \frac{\sum_{i = - m}^{m} \sum_{j = -
m}^{m} I \left ( x_{1} + i, x_{2} + j \right ) \omega \left ( {\bf
x, \xi} \right )}{\sum_{i = - m}^{m} \sum_{j = - m}^{m} \omega \left
( {\bf x, \xi} \right )},
\end{equation}
where ${\bf x} = \left ( x_{1}, x_{2} \right )^{T}$ is the and ${\bf
\xi} = \left ( \xi_{1}, \xi_{2} \right )^{T}$ are the center point
of the image window and the corresponding nearby point respectively,
$m$ is the window radius, $I = \left ( I_{\rm R}, I_{\rm G}, I_{\rm
B} \right )^{T}$ is the RGB color of a pixel. If we use Gaussian
filter, the weight function is:
\begin{equation} \label{weight_function}
\omega \left ( {\bf x, \xi} \right ) = c \left ( {\bf \xi, x} \right
) s \left ( I \left ( {\bf \xi} \right ), I \left ( {\bf x} \right )
\right ) = \exp \left ( - \frac{1}{2} \frac{\left | {\bf \xi} - {\bf
x} \right |^{2}}{{\sigma_{d}}^{2}}\right ) \exp \left ( -
\frac{1}{2} \frac{\left | I \left ( {\bf \xi} \right ) - I \left (
{\bf x} \right ) \right |^{2}}{{\sigma_{r}}^{2}}\right ),
\end{equation}
where the closer function $c \left ( {\bf \xi, x} \right )$ is
relative to the spatial distance $\left | {\bf \xi} - {\bf x} \right
|$ and the similar function $s \left ( I \left ( {\bf \xi} \right ),
I \left ( {\bf x} \right ) \right )$ is relative to the range
difference $\left | I \left ( {\bf \xi} \right ) - I \left ( {\bf x}
\right ) \right |$.

Bilateral filter method is an advanced method for noise removal
which aim at preserving the signal details. But on edges with
similar color the filter will be disabled as the weight function is
equal to one. So some more effective filter methods are needed to
solve this problem.

\section{Improved filter}

\subsection{Improved filter}

\noindent

Bilateral filter is a nonlinear filter which consider both the close
spatial distance and the similar colors, so it can smooth image with
edges preserved. But in color similar edge region, such as the
grasslands around the airport become yellow in autumn and winter or
some highlight disturbs exist, the color of the runway is similar
with the background,the spatial distance and the color difference
are both small, then the closer function $c \left ( {\bf \xi, x}
\right )$ and the similar function $s \left ( I \left ( {\bf \xi}
\right ), I \left ( {\bf x} \right ) \right )$ close to one
together, so the bilateral filter changed to:
\begin{equation}
I' \left ( {\bf x} \right ) \approx \frac{1}{\left ( 2 m + 1 \right
) \left ( 2 m + 1 \right )} \sum_{i = - m}^{m} \sum_{j = - m}^{m} I
\left ( x_{1} + i, x_{2} + j \right ).
\end{equation}

It is standard average filter that will cause blurring problem on
edges. More features are needed to add in filter so that we can
obtain robust and accurate filtered image.zy add

Expand the weight function $\omega \left ( {\bf \xi, x} \right )$ by
introducing an image texture similar related function $t \left (
{\bf \xi, x} \right )$, we have:
\begin{eqnarray}
\omega \left ( {\bf \xi, x} \right ) & = & c \left ( {\bf \xi, x}
\right ) s \left ( I \left ( {\bf \xi} \right ), I \left ( {\bf x}
\right ) \right ) t \left ( {\bf \xi, x} \right ) \nonumber\\
& = & \exp \left ( - \frac{1}{2} \frac{\left | {\bf \xi} - {\bf x}
\right |^{2}}{{\sigma_{d}}^{2}}\right ) \exp \left ( - \frac{1}{2}
\frac{\left | I \left ( {\bf \xi} \right ) - I \left ( {\bf x}
\right ) \right |^{2}}{{\sigma_{r}}^{2}}\right ) \exp \left ( -
\frac{1}{2} \frac{\left | T \left ( {\bf \xi} \right ) - T \left (
{\bf x} \right ) \right |^{2}}{{\sigma_{t}}^{2}}\right ),
\end{eqnarray}
here $t \left ( {\bf \xi, x} \right )$ denotes the closer degree of
texture type between pixels. We use the pixel with similar spatial,
color and texture values instead of core pixel, then the filtering
result can be improved. In the slick area, the difference between
neighboring pixels' value is small, the weight function almost equal
to $1$. The Improved filtering approximately correspond to a
standard mean filter; on the image edges the weight function
influence the weight of the nearby pixels, more similar pixels have
bigger weight and influence core pixel more. If the edge has similar
color but different texture features, the disturb from the pixel
with big texture difference can be rejected by $t \left ( {\bf \xi,
x} \right )$, so the image edge can be kept. The visualized
comparison and quantitative analysis filtering results of gray scale
and color image show in fig. 3 to fig. 5 respectively. Obviously,
texture similar function reinforce the edge of filtered image.

\subsection{Texture feature extraction}

\noindent

 Steerable filter decomposition (SFD) \cite{FA,CPMR,SW} is used to extract
texture features. SFD provides a finer frequency decomposition that
is more closely corresponds to human visual processing. The sub-band
of any orient can be linearly composed by a group of base filters,
so this method is flexible and operable. In this paper, SFD is used
to classify 6 texture types, and that of other orient can be
extended easily. Because texture is independent with color, we only
use grey scale image in this section.

Using a one level Steerable filter to decomposite image at an
arbitrary orientation $\theta$, the filtering sub-band is:
\begin{equation}
I^{\theta}_{1} = \cos \left ( \theta \right ) G_{1}^{0^{\circ}}
\star I + \sin \left ( \theta \right ) G_{1}^{90^{\circ}} \star I,
\end{equation}
where $I$ is the input image, $G$ is the kernel function and
\begin{equation}
G_{1}^{0^{\circ}} = \frac{\partial G}{\partial x}, \quad
G_{1}^{90^{\circ}} = \frac{\partial G}{\partial y}.
\end{equation}

Decomposing image on 4 orientation: $0^{\circ}$, $90^{\circ}$,
$45^{\circ}$ and $- 45^{\circ}$, we can obtain filtering
decomposition sub-band parameter $P_{j, 1}^{\theta}$. The median
local energy of core pixel on orientation $\theta$ is:
\begin{equation}
E^{\theta} = \frac{1}{n} \sum_{j = 1}^{n} \left ( P_{j, 1}^{\theta}
\right )^{2}.
\end{equation}
For each pixel, a four dimension median local energy vector can be
obtained.

\begin{figure}
 \centering
 \subfigure[Input Image]{
  \label{texture:subfig:a}
  \includegraphics[width=4cm]{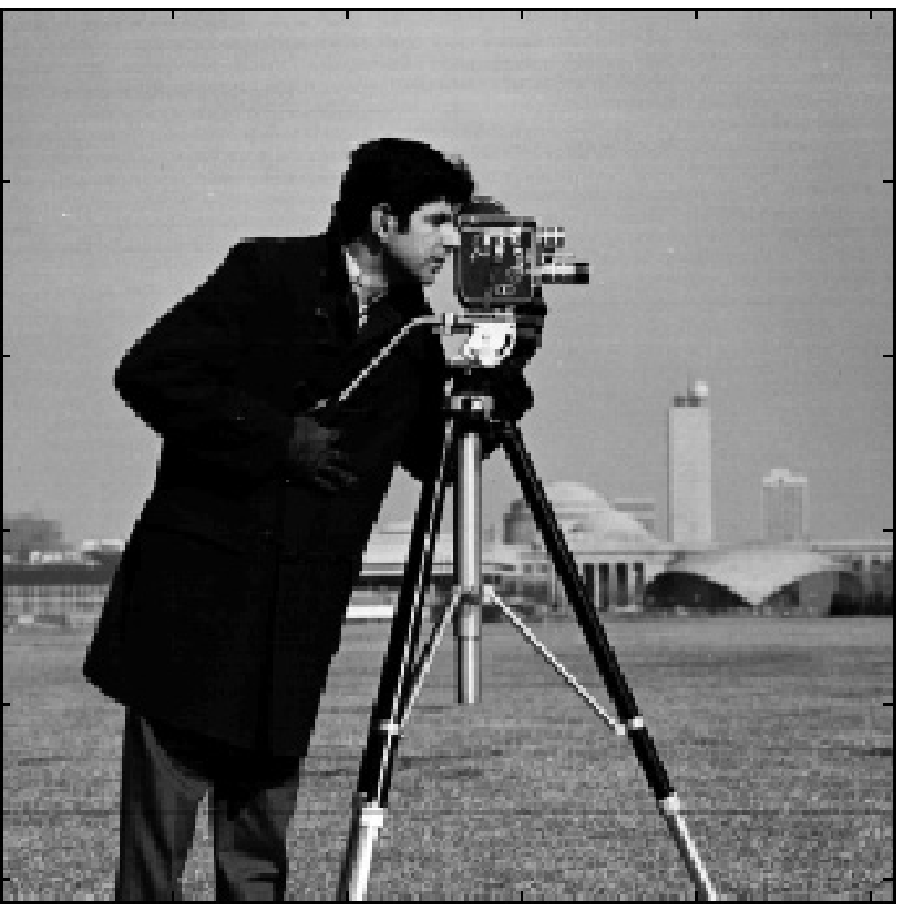}}
 \hspace{1cm}
 \subfigure[Texture Image]{
  \label{texture:subfig:b}
  \includegraphics[width=4cm]{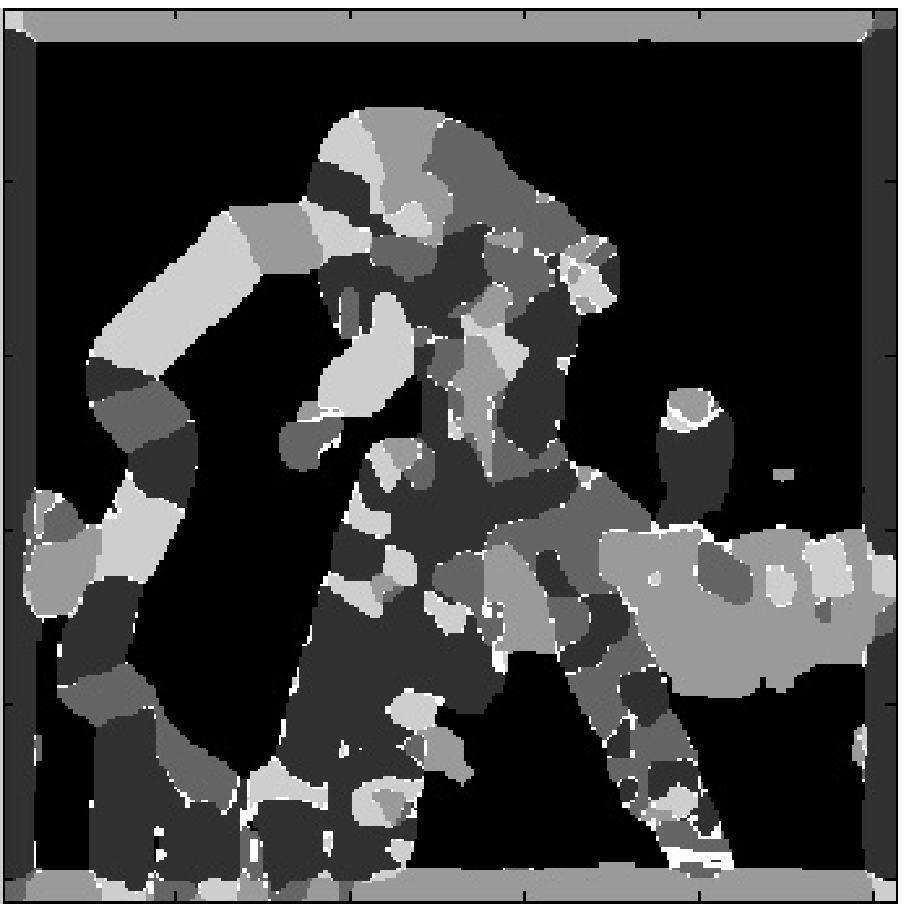}}
 \caption{Texture Feature Image}
 \label{texture:subfig}
\end{figure}

A pixel in texture region can be classified into one of six texture
categories based on $E^{\theta}$. Firstly, orientation with biggest
energy is texture orientation; but if the energies on all
orientations is small enough (smaller than giving threshold
$T_{1}$), then there isn't enough energy of all sub-band, the
texture is regarded as smooth; if at least there are two energies
are big enough and close to each other, then there isn't dominant
energy, texture is complex. Classifying texture of all pixels, we
can obtain a texture matrix. An input image and its classification
result are illustrated in Fig.\ref{texture:subfig}, in the
figuration colors with different grey scales denote different
textures.

\subsection{Parameter selection}

\noindent

\begin{figure}
 \centering
 \includegraphics[width=3cm]{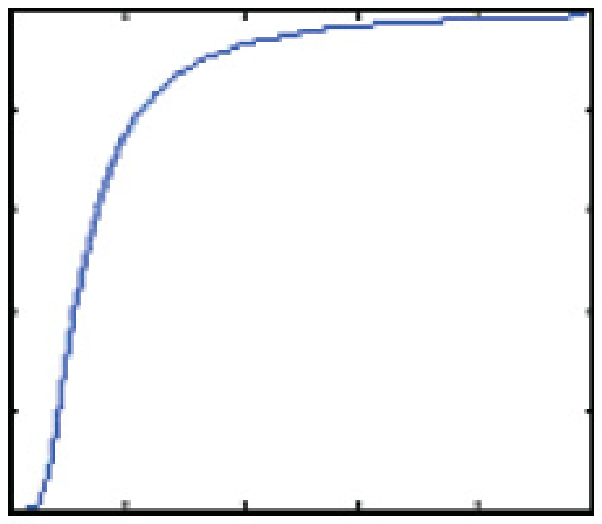}
 \caption{relationship of $t \left ( {\bf \xi, x} \right )$ and
 $\sigma_{t}$}
 \label{relation}
\end{figure}

The relationship of $\sigma_{t}$ and the weight function $t \left (
{\bf \xi, x} \right )$ is showed in Fig.\ref{relation}, where $x$
and $y$ axes represent $\sigma_{t}$ and $t \left ( {\bf \xi, x}
\right )$ respectively. When $\sigma_{t} \rightarrow \infty$, the
weight function $t \left ( {\bf \xi, x} \right ) \rightarrow 1$,
improved filter approximately equal to standard bilateral filter; if
$\sigma_{t}$ is very small, $t \left ( {\bf \xi, x} \right )$
becomes more important in $\omega \left ( {\bf x, \xi} \right )$,
filtered image relies on texture mainly.

\section{Simulation results and applications in runway image}

\subsection{Simulation results}

\noindent

\begin{figure}
 \centering
 \subfigure[Input Image added noise]{
  \label{noise:subfig:a}
  \includegraphics[width=4cm]{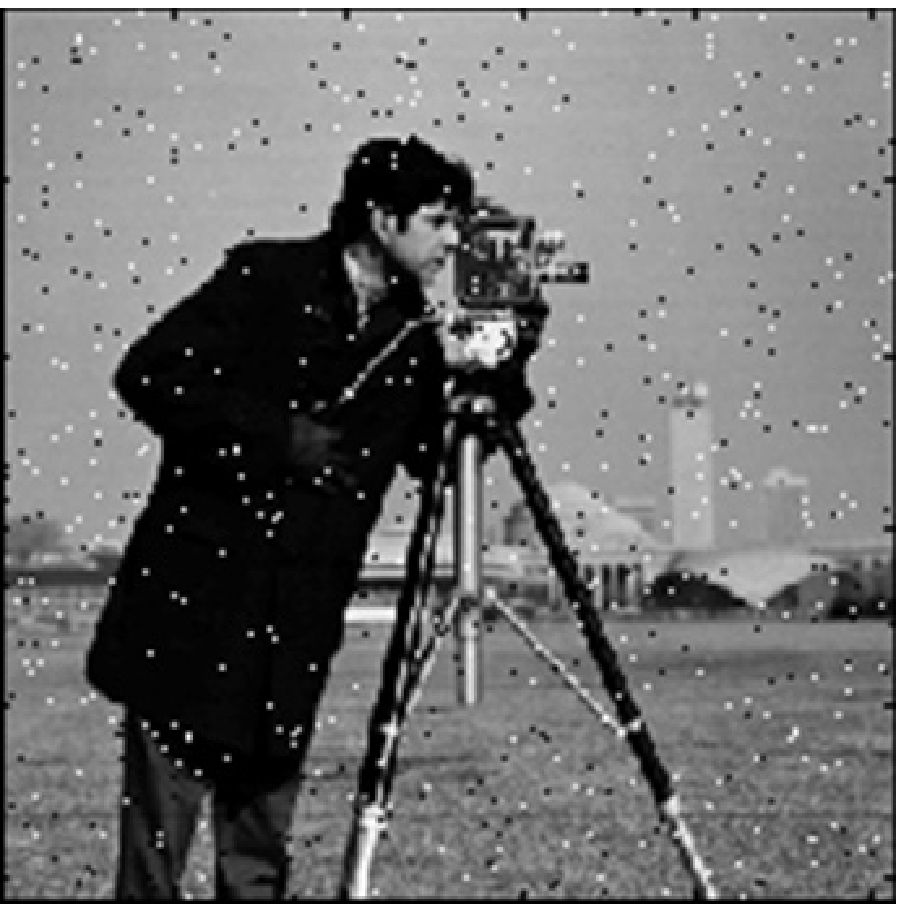}}
 \hspace{3mm}
 \subfigure[bilateral filtering result]{
  \label{noise:subfig:b}
  \includegraphics[width=4cm]{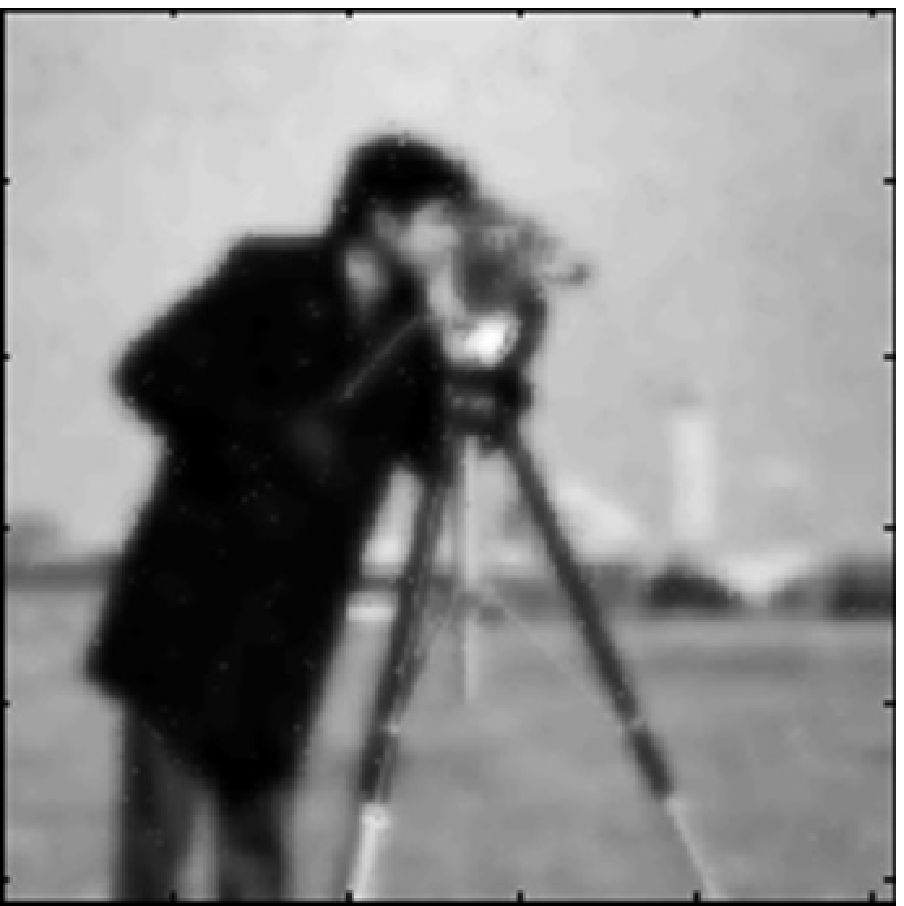}}
 \hspace{3mm}
 \subfigure[Multilateral filtering result]{
  \label{noise:subfig:b}
  \includegraphics[width=4cm]{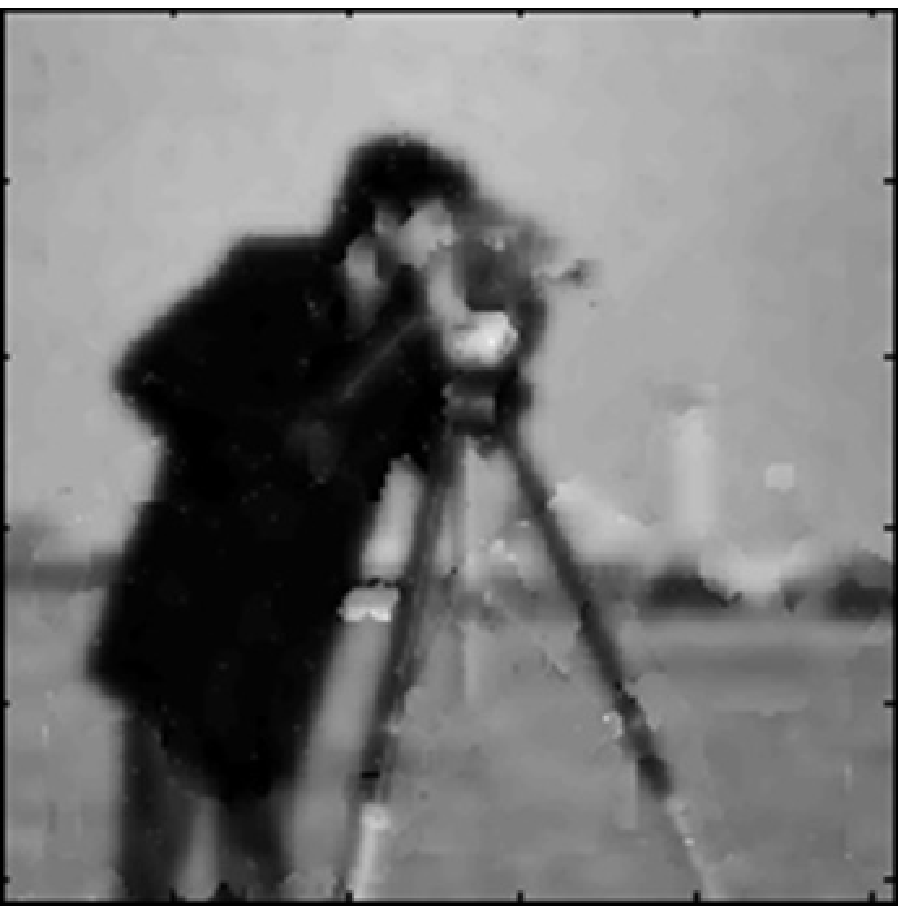}}
 \caption{Filtering result comparison of grey scale image}
 \label{noise:subfig}
\end{figure}

\begin{figure}
 \centering
 \subfigure[Input Image]{
  \label{tower:subfig:a}
  \includegraphics[width=3cm]{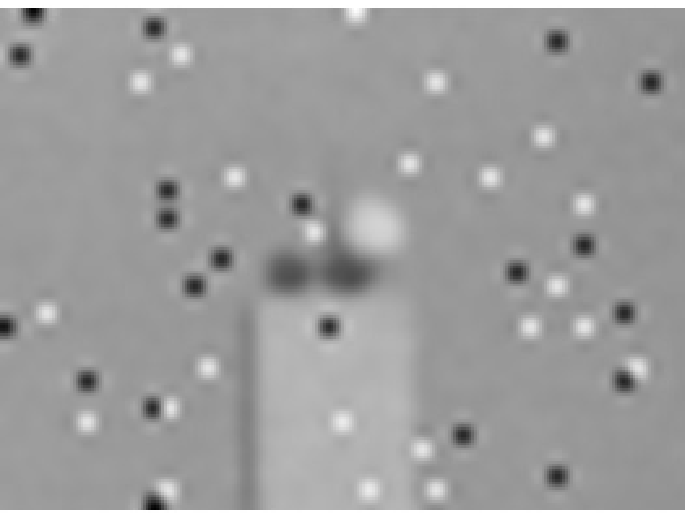}}
 \hspace{3mm}
 \subfigure[bilateral filtering result]{
  \label{tower:subfig:b}
  \includegraphics[width=3.1cm]{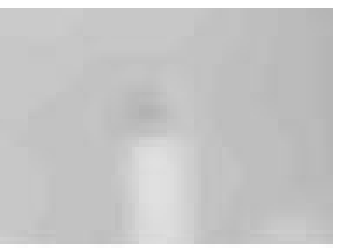}}
 \hspace{3.0mm}
 \subfigure[Multilateral filtering result]{
  \label{tower:subfig:b}
  \includegraphics[width=3cm]{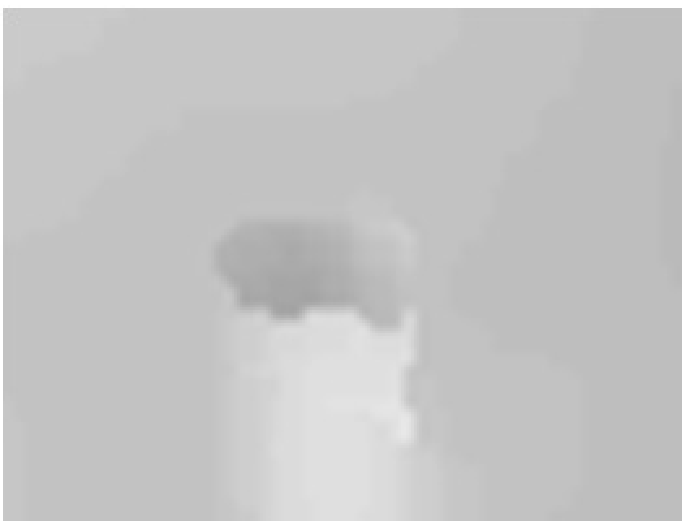}}
 \caption{Detail comparison of fig.3}
 \label{tower:subfig}
\end{figure}

Fig.\ref{noise:subfig} shows the filtering results for a cameraman
image added salt and pepper noise after two times filtering.
Visualized comparison shows that there is not obviously improvements
on edges with different colors. But with similar color, the
multilateral filter can keep the image edge detail and have
obviously advantage over BF, this result accord with academic
analysis. Fig.\ref{tower:subfig} illustrates the amplificatory tower
top comparing result. In the input image, the colors of the tower
top are closer to the sky. the tower top can not easily be
identified in the bilateral filtered image but it is obviously in
the multilateral filtered image.

In order to analyze the performance of filters, here we calculate
the signal noise ratio (SNR) and the image edge retain exponent
$E_{P}$ of the filtered images. The SNR is defined as the ratio of
the grey scale value mean square of the input image and the
difference of the grey scale value mean square between the input and
the filtered image. The image edge retain exponent $E_{P}$ denotes
the edge preserving ability of the filters for the image after
filtering. The bigger value of $E_{P}$ means the better edge
preserving ability. The expression of $E_{P}$ is
\begin{equation}
E_{P} = \frac{\sum_{i = 1}^{m} \left | G_{R_{1}} - G_{R_{2}} \right
|_{\rm filterred}}{\sum_{i = 1}^{m} \left | G_{R_{1}} - G_{R_{2}}
\right |_{\rm input}},
\end{equation}
here $m$ is the number of samples, $G_{R_{1}}$ and $G_{R_{2}}$ are
the grey scale value of neighbor pixels on the image edge of grey
scale image respectively.Color image will be converted to grey scale
image to calculate $E_{P}$.

Tab.\ref{Tab_Para_comp} shows the statistic dada of the parameter
comparison. Here the SNR and the Edge preserving exponent $E_{P}$
are calculated using the whole image. From Tab.\ref{Tab_Para_comp}
we can conclude that the SNR and the edge retaining ability of the
improved filter exceed the standard bilateral filter, both on
horizontal and vertical orientation.

\begin{table}
\caption{\label{Tab_Para_comp} Parameter comparison of the bilateral
and multilateral filtering results Image added noise}
 \centering
\begin{tabular}{|c|c|c|c|c|c|c|}
\hline & \multicolumn{2}{|c|}{} & \multicolumn{4}{|c|}{Edge
preserving exponent $E_{P}$}
\\
\cline{4-7} Image added noise &
\multicolumn{2}{|c|}{\raisebox{1.3ex}[0pt]{SNR}} &
\multicolumn{2}{|c|}{horizontal} & \multicolumn{2}{|c|}{vertical}
\\
\cline{2-7} & {\footnotesize Bilateral} & {\footnotesize
multilateral} & {\footnotesize Bilateral} & {\footnotesize
multilateral} & {\footnotesize Bilateral} & {\footnotesize
multilateral} \\
\hline Cameraman salt & & & & & & \\
and pepper's noise & \raisebox{1.3ex}[0pt]{14.1457} &
\raisebox{1.3ex}[0pt]{14.3024} & \raisebox{1.3ex}[0pt]{0.1259} &
\raisebox{1.3ex}[0pt]{0.1821} & \raisebox{1.3ex}[0pt]{0.3665} &
\raisebox{1.3ex}[0pt]{0.4464} \\
\hline Cameraman & & & & & & \\
Gaussian noise & \raisebox{1.3ex}[0pt]{12.7664} &
\raisebox{1.3ex}[0pt]{12.8754} & \raisebox{1.3ex}[0pt]{0.1447} &
\raisebox{1.3ex}[0pt]{0.2339} & \raisebox{1.3ex}[0pt]{0.5459} &
\raisebox{1.3ex}[0pt]{0.6697} \\
\hline Lena salt & & & & & & \\
and pepper's noise & \raisebox{1.3ex}[0pt]{13.4620} &
\raisebox{1.3ex}[0pt]{13.5794} & \raisebox{1.3ex}[0pt]{0.0716} &
\raisebox{1.3ex}[0pt]{0.1200} & \raisebox{1.3ex}[0pt]{0.3290} &
\raisebox{1.3ex}[0pt]{0.4207} \\
\hline Lena & & & & & & \\
Gaussian noise & \raisebox{1.3ex}[0pt]{12.0803} &
\raisebox{1.3ex}[0pt]{12.1530} & \raisebox{1.3ex}[0pt]{0.1032} &
\raisebox{1.3ex}[0pt]{0.1851} & \raisebox{1.3ex}[0pt]{0.5410} &
\raisebox{1.3ex}[0pt]{0.6798} \\
\hline Pears (color image) salt & & & & & & \\
and pepper's noise & \raisebox{1.3ex}[0pt]{17.5300} &
\raisebox{1.3ex}[0pt]{17.6996} & \raisebox{1.3ex}[0pt]{0.1131} &
\raisebox{1.3ex}[0pt]{0.1761} & \raisebox{1.3ex}[0pt]{0.3709} &
\raisebox{1.3ex}[0pt]{0.4514} \\
\hline Pears (color image) & & & & & & \\
Gaussian noise & \raisebox{1.3ex}[0pt]{16.9481} &
\raisebox{1.3ex}[0pt]{17.1050} & \raisebox{1.3ex}[0pt]{0.1514} &
\raisebox{1.3ex}[0pt]{0.2266} & \raisebox{1.3ex}[0pt]{0.6104} &
\raisebox{1.3ex}[0pt]{0.7000} \\ \hline
\end{tabular}
\end{table}

In order to compare edge preserve ability of two kinds of filters in
different strength of noise, here we give the simulation of Lena
added salt and pepper noise. The results of $E_{p \rm Multi} / E_{p
\rm Bi}$ are shown in Tab.\ref{Tab_Noise_cofficent_comp}, the
horizontal and vertical $E_{P}$ comparisons are also included. Here
$E_{p \rm Multi}$ and $E_{p \rm Bi}$ represent the edge preserve
parameters of multilateral and bilateral filters respectively. From
the analysis of Tab.\ref{Tab_Noise_cofficent_comp}, the increase of
the noise coefficients can cause an obvious increase of the ratio of
the edge preserve parameters of the multilateral filter and
bilateral filter. Compared with bilateral filter, this indicates
that in a more serious noise condition, the multilateral filter has
a stronger edge preserving ability.

\begin{table}
\caption{\label{Tab_Noise_cofficent_comp} Comparison of edge
preserve ability in different noise conditions}
 \centering
\begin{tabular}{|c|c|c|} \hline
Noise coefficient & horizontal & vertical \\
\hline 0.01 & 1.2992 & 1.2739 \\
\hline 0.03 & 1.5758 & 1.2661 \\
\hline 0.05 & 2.0938 & 1.2753 \\
\hline 0.07 & 2.3385 & 1.3668 \\
\hline
\end{tabular}
\end{table}

\begin{table}
\caption{\label{Tab_Epara_comp} Filtering parameter comparison of
different $\sigma_{t}$} \centering
\begin{tabular}{|c|c|c|c|}
\hline & \multicolumn{2}{|c|}{Edge preserving exponent $E_{P}$} & \\
\cline{2-3}
\raisebox{1.3ex}[0pt]{$\sigma_{t}$} & horizontal & vertical &
\raisebox{1.3ex}[0pt]{SNR} \\
\hline 0.1 & 0.0340 & 0.2776 &           12.9826 \\
\hline 1 & 0.0214 & 0.2616 & 12.9527 \\
\hline 10 & 0.0150 & 0.2339 & 12.8309 \\
\hline 100 & 0.0150 & 0.2341 & 12.8284 \\
\hline
\end{tabular}
\end{table}

Tab.\ref{Tab_Epara_comp} gives the comparison of filtering parameter
with different $\sigma_{t}$.  As $\sigma_{t}$ increase, SNR decrease
gradually and the edge preserving parameter has an obvious decrease
trend. The simulation results show that as $\sigma_{t} \rightarrow
\infty$, SNR and $E_{p}$ tend to the values of bilateral filter.
When $\sigma_{t} < 0.1$, SNR and $E_{p}$ will go to fixed. This
result is consistent with the theoretical analysis.

\subsection{Airport runway image filtering}

\noindent

The key of the UAV autonomous landing problem is accurate
recognition of the airport runway. However because there are amount
of inevitable noise and some disturbs in adopted image sequences,
such as the runway marks and the plane shadows, then the accuracy of
the runway recognition is effected. So filtering is necessary. When
runway and background colors are similar, the effect of standard
bilateral filter is not good.  In this subsection we / use an
improved filtering method to process airport runway image.

\begin{figure}
 \centering
 \subfigure[Input image and its histogram]{
  \label{Airport:subfig:a}
  \includegraphics[width=3.5cm]{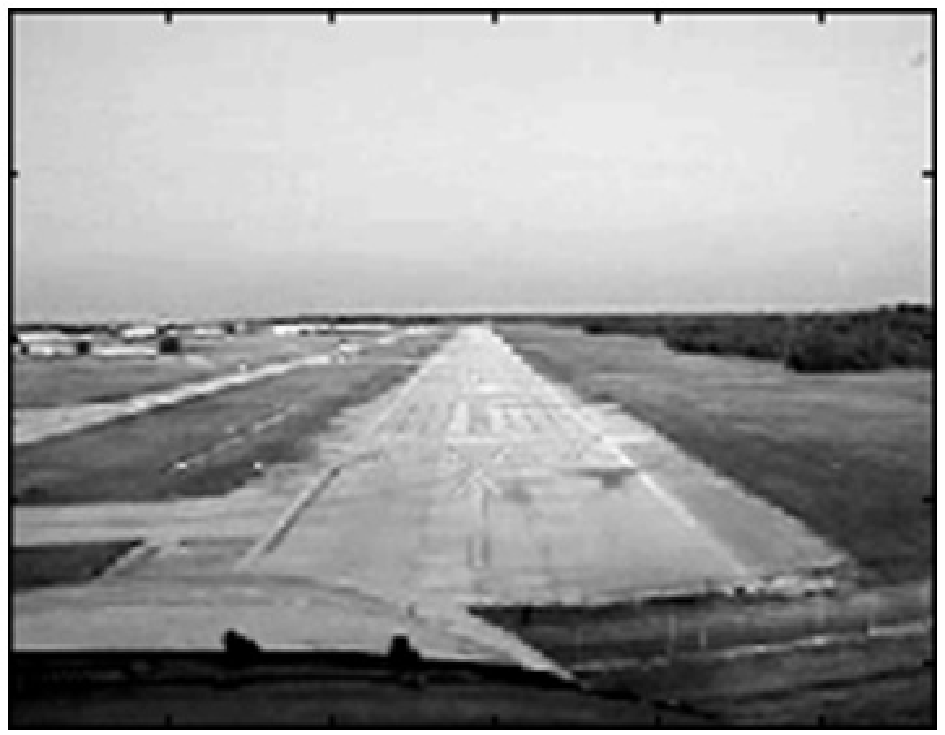}
  \includegraphics[width=3.5cm]{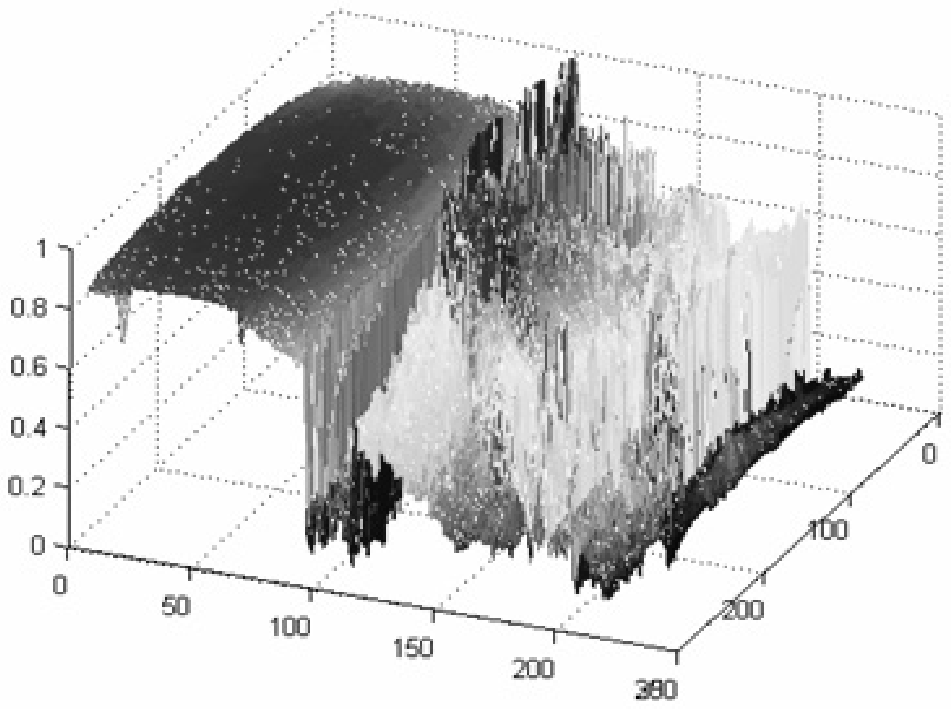}}
 \hspace{3mm}
 \subfigure[Filtered image and its histogram ]{
  \label{Airport:subfig:c}
  \includegraphics[width=3.5cm]{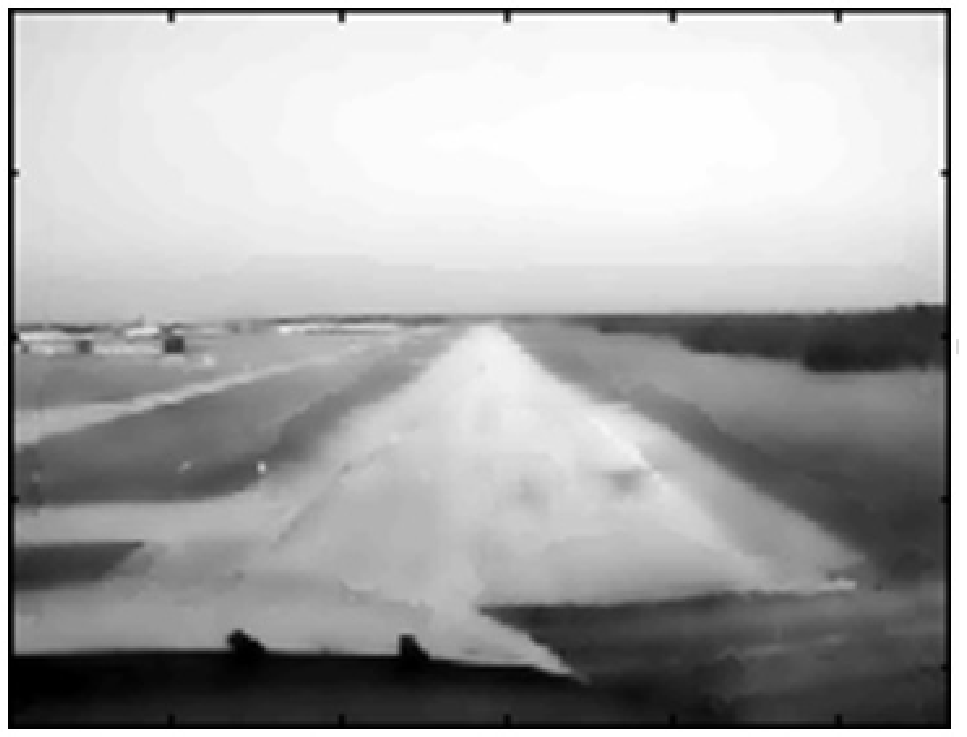}
  \includegraphics[width=3.5cm]{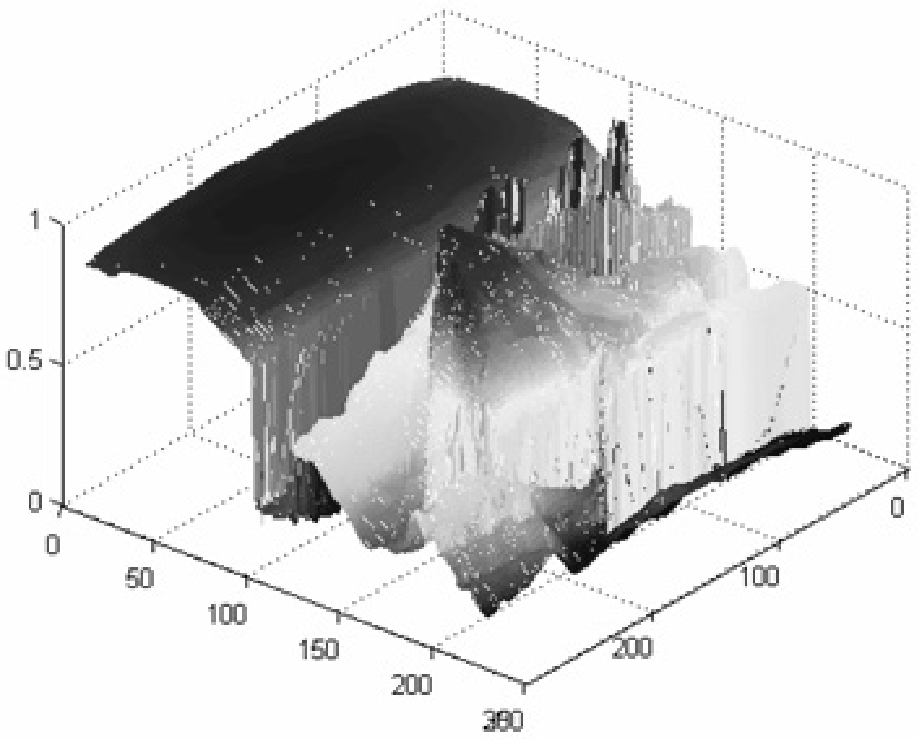}}
 \caption{Airport image filtering}
 \label{Airport:subfig}
\end{figure}

\begin{table}
\centering \caption{\label{Tab_SNR_comp} Parameter comparison of
bilateral and multilateral filters SNR Edge preserving exponent
$E_{p}$}
\begin{tabular}{|c|c|c|c|c|c|}
\hline \multicolumn{2}{|c|}{} & \multicolumn{4}{|c|}{Edge preserving
exponent $E_{P}$}
\\
\cline{3-6} \multicolumn{2}{|c|}{\raisebox{1.3ex}[0pt]{SNR}} &
\multicolumn{2}{|c|}{horizontal} & \multicolumn{2}{|c|}{vertical}
\\
\hline Bilateral & multilateral & Bilateral & multilateral &
Bilateral & multilateral \\
\hline 29.4747 & 29.6184 & 0.3275 & 0.3675 & 0.5219 & 0.5643 \\
\hline
\end{tabular}
\end{table}

Fig.\ref{Airport:subfig} illustrate the filtering results of the
airport runway image. From the figure, we can see that the runway
information is preserved while the noise is depressed in the
filtered image, the mark and shadow on the runway is restrained. It
is benefit to segment and recognition. Especially when the grass
becomes yellow in autumn and winter or there are highlight disturbs,
the colors of runway and background are similar, at this time the
standard bilateral filter doesn't work well, but the multilateral is
suitable. Tab.\ref{Tab_SNR_comp} gives a quantitatively comparison
of the bilateral filtering result and the improved filter result for
the input image of fig. 6. We find that the SNR and the edge
preserving exponent of the improved filter are better in actual
situation .

Here we considered not only the spatial and color closer degree, but
also the texture similarity, so the filter performance is improved.
The Multilateral filter is simple and non-iterative as the standard
bilateral filter, but more robust. The application of the improved
filter in airport image processing is effective.

\section{Conclusion}

\noindent

By considering the features of the airport runway image filtering,
in this paper we proposed an improved bilateral filtering method.
Firstly we used the local median energy of steerable filtering
decomposition to extract the texture feature of the image, then we
gave out the filtered image by using the texture similar, spatial
closer and color similar functions. The effect of the weighting
function parameters is qualitatively analyzed also. Both the
visualized comparisons and the quantitative analysis of the
filtering results for the noise added grey scale and the color image
showed that the improved method is more effective than the standard
bilateral filtering method to preserve the image edge. The advantage
of the improved filter is obvious as the noise strength increasing.
The filtering results of the airport runway image showed that the
improved filtering method can remove noises and disturbs
effectively, and has better SNR and edge preserving ability.

While the multilateral filter relies on the texture extraction
result. An unappropriate texture feature will yield disturb and
influence the filtering result, so before filtering, it is very
important to obtain an accurate and appropriate texture feature.

\end{document}